\documentclass[conference]{IEEEtran}
\IEEEoverridecommandlockouts
\usepackage{cite}
\usepackage{amsmath,amssymb,amsfonts}
\usepackage{algorithmic}
\usepackage{graphicx}
\usepackage{textcomp}
\usepackage{xcolor}
\usepackage{geometry}
\def\BibTeX{{\rm B\kern-.05em{\sc i\kern-.025em b}\kern-.08em
    T\kern-.1667em\lower.7ex\hbox{E}\kern-.125emX}}
\begin{document}

\title{Using Machine Learning to Evaluate Real Estate Prices Using Location Big Data\\
}

\author{\IEEEauthorblockN{Walter Coleman, Ben Johann, Nicholas Pasternak, Jaya Vellayan, Natasha Foutz, and Heman Shakeri}
\IEEEauthorblockA{\textit{University of Virginia, txx3ej, bmj7sf, nfp5ga, jlv7cr, ynf8a, and hs9hd@virginia.edu}
}}

\maketitle

\begin{abstract}
With everyone trying to enter the real estate market nowadays, knowing the proper valuations for residential and commercial properties has become crucial. Past researchers have been known to utilize static real estate data (e.g. number of beds, baths, square footage) or even a combination of real estate and demographic information to predict property prices. In this investigation, we attempted to improve upon past research. So we decided to explore a unique approach – we wanted to determine if mobile location data could be used to improve the predictive power of popular regression and tree-based models. 
To prepare our data for our models, we processed the mobility data by attaching it to individual properties from the real estate data that aggregated users within 500 meters of the property for each day of the week. We removed people that lived within 500 meters of each property, so each property's aggregated mobility data only contained non-resident census features. On top of these dynamic census features, we also included static census features, including the number of people in the area, the average proportion of people commuting, and the number of residents in the area. Finally, we tested multiple models to predict real estate prices. Our proposed model is two stacked random forest modules combined using a ridge regression that uses the random forest outputs as predictors. The first random forest model used static features only and the second random forest model used dynamic features only. Comparing our models with and without the dynamic mobile location features concludes the model with dynamic mobile location features achieves 3\% lower mean squared error than the same model but without dynamic mobile location features.

\end{abstract}

\section{Introduction}
Real estate pricing is often qualitative, and realtors consider a variety of real estate-based features such as location, number of bedrooms, number of bathrooms, square footage, etc., that influence how a given residential or commercial property is priced. Real estate pricing is not necessarily consistent among different property owners or realtors; each realtor or property owner has different techniques for pricing real estate. This makes the problem of predicting real estate prices to be complicated. People place different values on various property features, so it is challenging to create a model that accurately reflects this human decision-making process. To narrow down this problem, we focused on the real estate market in Washington D.C. and the surrounding neighborhoods. Cities, including Washington D.C., tend to have more real estate and higher prices than rural areas. Usually, real estate prices are dependent on the physical features of a home and the common demographic characteristics associated with the home’s location. However, our objective is to determine if features engineered from mobile location data increase the accuracy of our models.

\subsection{Project Goal}
This project aims to determine if the inclusion of mobile location data, on top of typical real estate and demographic features, will improve our ability to predict real estate prices accurately. However, we noticed minimal research on modeling real estate prices using mobile location data when researching similar projects. Hence, it is paramount to discover the benefit or lack thereof of using mobile location data in determining real estate prices. Additionally, there was a lack of research on the pricing of commercial properties. So through carrying out this investigation, we also want to construct a solid commercial real estate pricing model and see how the mobility data impacts the commercial side compared to the residential.

\subsection{Research Questions}

\begin{itemize}
  \item Past real estate pricing research has only considered static real estate and demographic data to build prediction models. Thus our primary objective is to determine if mobile location data can be leveraged to improve real estate pricing models?
  \item In carrying out this primary objective, we also want to compare the impact of using mobility features to predict the prices of commercial and residential properties?
\end{itemize}

\section{Literature Review}
Before starting our project, we researched other journals that worked with real estate pricing and/or mobile location data. Through all this research, we understood better how to approach our specific research questions. We determined that finding studies that used location data to predict real estate prices was challenging. We also could not find any research on the pricing of commercial properties. We mainly came across residential real estate studies that utilized real estate property and demographic data to predict prices and separate studies that used location data for non-real estate-related reasons. 

\subsection{Related Work}
Before conducting our experiments, we reviewed roughly thirty journals on predicting real estate pricing using census data and location big data. Authors in [1] used census data better to understand the effects of school quality on neighborhood demand. Additionally, they discussed self-segregation based on neighbor race and education. Finally, the authors developed a framework for estimating household preferences for school and neighborhood attributes. This paper was helpful for feature selection and engineering, specifically when we considered what census-based mobile location features to create and keep in our model. In our model, we used this prior research to ensure we kept influential census features, like race and education levels.

Another journal we considered is Understanding Urban Gentrification through Machine Learning [2]. These authors used machine learning techniques to predict neighborhood improvement, which can lead to gentrification. This journal article was especially helpful after we finished our modeling because it gives us insights into how to look for gentrification, which was helpful as we tried to get a better understanding of social disparity. 

The journal article Artificial Intelligence for Modeling Real Estate Price Using Call Detail Records and Hybrid Machine Learning Approach, is one of the more similar experiments to ours. The authors use Call Detail Records (CDR) to track mobility and price real estate using machine learning [3]. Then, they find the home and work locations of each person. We also determined the home and work locations of each person. Then, they determined the home-work distance, entropy, radius of gyration are calculated for every SIM card. Finally, they determined the average real estate price for every cell via polygons. This was helpful for giving us insight into how to use the determined home and work locations effectively. 

Additionally, The Economic Value of Neighborhoods: Predicting Real Estate Prices from the Urban Environment was helpful because it used machine learning to infer how much a neighborhood affects sales price using many different neighborhood features to improve the prediction of home prices [4]. We were able to utilize this information to create and keep neighborhood based features that would improve the accuracy of our models. 

The Effect of Home-Sharing on House Prices and Rents: Evidence from Airbnb, was a very interesting and insightful journal article. It used AirBnB data to analyze  how home-sharing impacted real estate pricing [5]. They found that home sharing has increased real estate. This was something that we would not have considered in the analysis of our results without having read this article. 

Another eye opening article is The Effects of Rent Control Expansion on Tenants, Landlords, and Inequality: Evidence from San Francisco. The authors discuss the impact of rent control on gentrification and housing mobility [6]. This study was  useful in understanding the impact of the rent control laws in DC. Moving more into modeling, Ridge Estimators in House Valuation Models was useful because it discusses the use of Ridge estimators for evaluating single family residencies in the presence of multicollinearity. Since ordinarily least squares regression typically exhibits multicollinearity, they decided to use ridge regression in their analysis. This article influenced us to choose ridge regression as one of the models in our study.

\section{Data Description}
We utilized three different types of data in this project: Mobile location data, census data, and real estate data. The mobile location data for this project was collected by a data aggregating company, across hundreds of commonly used mobile apps, from news, weather, map, to fitness, in strict compliance with privacy regulations including GDPR and CCPA. This allows the data to be more accurate than cell tower or social media tags data. The data includes both Android and iOS users. Each individual is tracked every five minutes or after they move more than 100 meters. The data records an anonymous id, timestamp, longitude, latitude, dwell time, speed, point-of-interest names, and platform id. We were able to compress that using the Python package Infostop into individual stops. Hence, instead of having multiple rows with one person in a location for longer than five minutes, we are able to represent each stop at a single location with one row with timestamps for start and stop time. If one user does not move more than 50 meters for 5 minutes or longer, we represent this as one data point instead of many. From here, home locations were estimated by finding the centroid of each user’s most frequent stop location from Tuesday until Friday between the hours of 9pm and 7am. With this information, census block group demographics were attached to individual users offering more insight into the areas in which they lived in. These census features included median household income, age, race, level of education, unemployment, and more.

We also had access to residential real estate data from Mashvisor. This data included over 110,000 properties from Washington D.C., Virginia, and Maryland that were listed in 2019. It had numerous property features such as square footage, bedrooms, bathrooms, and location (longitude and latitude). Once again census block group features were attached to this dataset. Some of these properties can be seen below colored by price (darker is more expensive in Fig. 1).

Finally, a separate real estate data set from the Washington D.C. and Arlington, VA tax offices was needed to answer our second research question since it had both commercial and residential properties. This data consisted of 250,000 total properties. However it had far fewer property features and only had property tax assessment (rather than list price). However, it enabled the comparison of commercial and residential properties in terms of mobility’s impact. These properties can be seen below with orange indicating commercial and blue indicating residential (Fig 2).

\begin{figure}[htp]
    \centering
    \includegraphics[width=6cm]{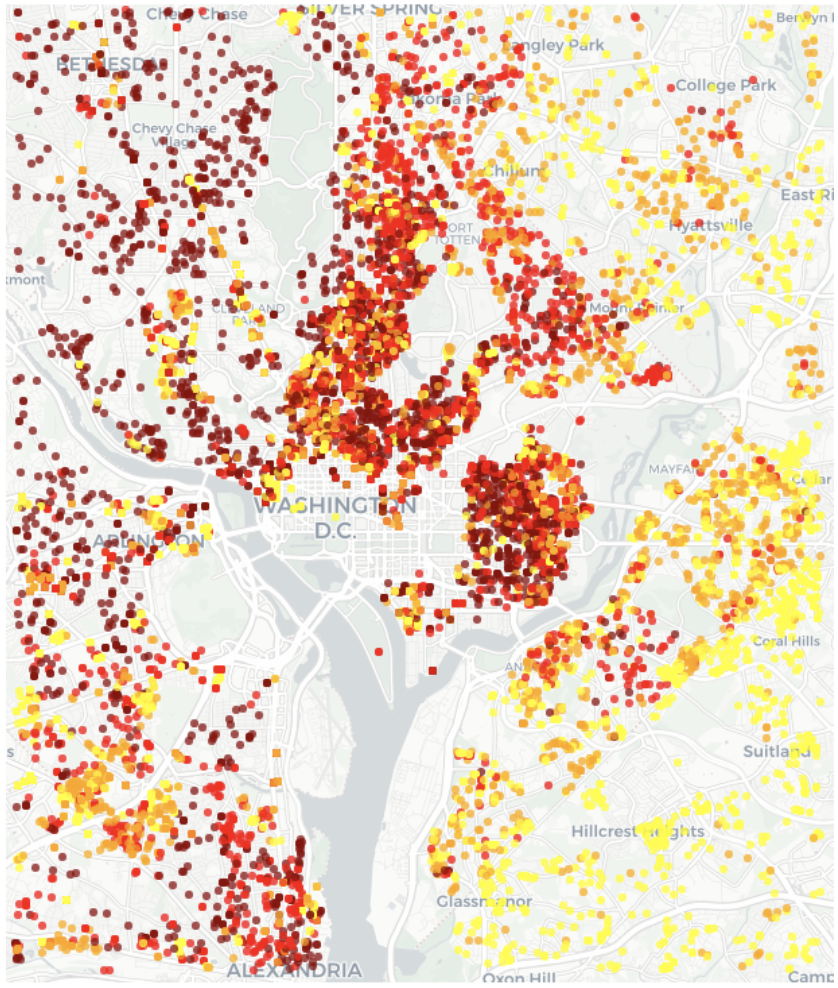}
    \caption{Mashvisor residential properties}
    \label{fig:galaxy}
\end{figure}

\begin{figure}[htp]
    \centering
    \includegraphics[width=6cm]{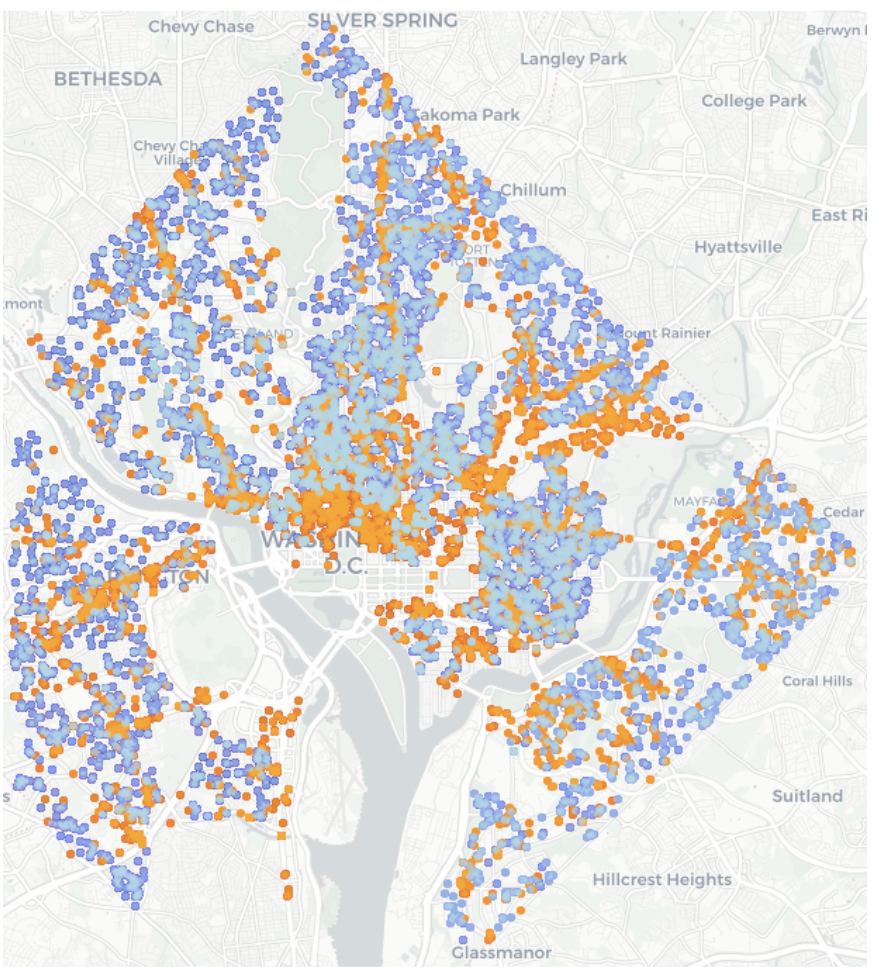}
    \caption{Commercial and Real Estate Properties}
    \label{fig:galaxy}
\end{figure}

\section{Methodology}
The processed mobility data was attached to individual properties from the real estate data by aggregating users that were within 500 meters of the property each day of the week. For each property all of these home location census features were averaged. These dynamic census features were complemented by static census features which were simply derived from the census block group where the property was located as mentioned above. Since the static features account for users that live in the same cbg as the property, residents were dropped from the census mobility aggregation (to avoid double counting). In other words, the dynamic census features only measure non-residents within 500 meters of a property on a given day of the week. 

In addition to dynamic census features, the total number of people in the area, the average proportion of people commuting, and the total number of residents in the area within 500 meters were calculated for each day of the week. Together the full feature space gives insight into mobility patterns surrounding a given property while also controlling for static census and property features which have extensively been shown to impact real estate pricing in past research. After connecting these data sources, there were over 12,000 properties ready for analysis (many properties were outside the bounds of the mobility data). 

Multiple machine learning techniques were applied to predict the list price of the properties. Drawing from previous research in the field, we tested different penalized regression and tree-based models. Our final model combined both methods. Two random forests were stacked. One model used the static features (property and static census) while the other model used only the dynamic mobile location data. These two models were combined using a ridge regression which used the predictions of both random forests as predictors. Random forest is a tree-based ensemble model that utilizes bootstrap aggregation, or bagging, to develop independent and individual regression models from random bootstrap samples of the training set. The predictions of the independent regression models are then averaged to produce the final output. It performs well when there are lots of interactions between variables (which we expected in our data). Ridge regression is a type of linear regression that utilizes a shrinkage estimator called a ridge estimator that works to shrink model coefficients, but not all the way to zero to regularize the model. This form of regression performs L2 regularization, which introduces a penalty term equal to the square of the magnitude of the model coefficients. By stacking the two random forests, one model is able to specialize in finding the relationship between the static variables and price while the other can focus on the dynamic data. Furthermore, this approach enables each model to be tuned separately in order to increase performance. 

In addition to the model(s) mentioned above, we ran separate models using property tax data from the revenue offices of Washington, DC and Arlington, VA that we could separate by residential and commercial properties. These models only used the dynamic mobility data to make predictions. We considered condos, apartments, single family homes and townhouses to be ‘residential’ and offices, restaurants, hotels, vacant lots, schools and churches to be ‘commercial.’ Using the same join methods mentioned above, we sampled approximately 5000 records from each of the two categories and removed all properties valued at less than \$50,000. To improve R squared, we performed a log transformation on the response, the 2019 price assessments. We ran two separate random forest models each with 700 estimators and a ten percent split in training and test data to compare each model’s shapley values.

\section{Results}
The results of the stacked random forest regressor model indicate that mobile location data can improve the performance of residential real estate models when predicting list price. To evaluate the effectiveness of the mobile data, two ensemble models were compared. One ensemble incorporated both dynamic location data and static data as described above, while the other simply used the static features in both models within the ensemble. The model that incorporated the location data had a 3\% smaller MSE than the static model when evaluating on a holdout sample of over 1200 properties.

We also found the shapley values of the residential and commercial models (predicting tax assessment) in order to highlight any differences between mobility’s impact on these property types. The commercial random forest had an MSE of 1.77 and an R2 of 0.45, while the residential random forest had an MSE of 0.30 and an R2 of 0.48. The top 20 features are listed in descending order (most important to least) below. The number 0 corresponds to Monday and 6 to Sunday. All features aside from  “people in area” are averaged across all non-residents in the area for the given time.

\begin{figure}[htp]
    \centering
    \includegraphics[width=7cm]{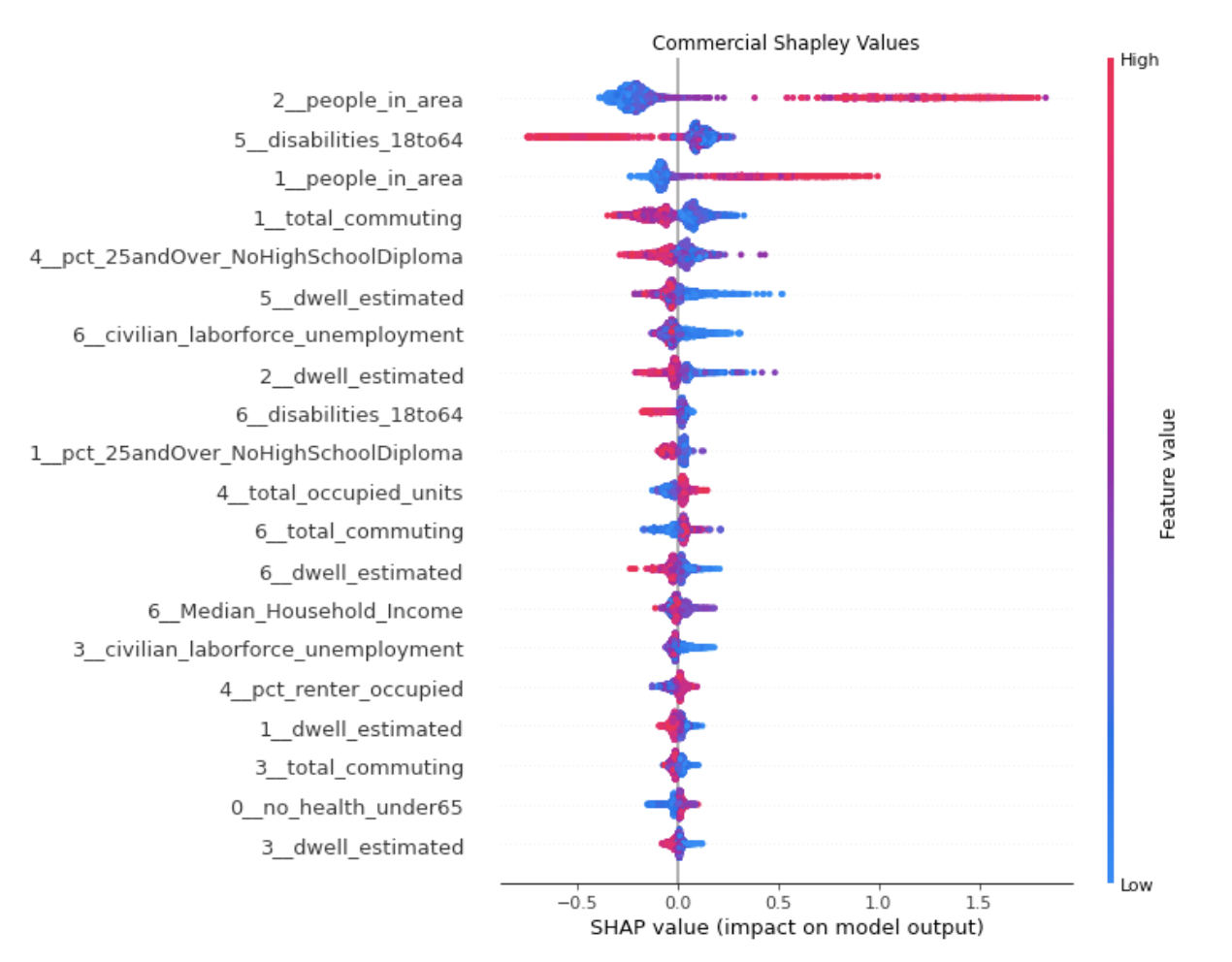}
\end{figure}

\begin{figure}[htp]
    \centering
    \includegraphics[width=7cm]{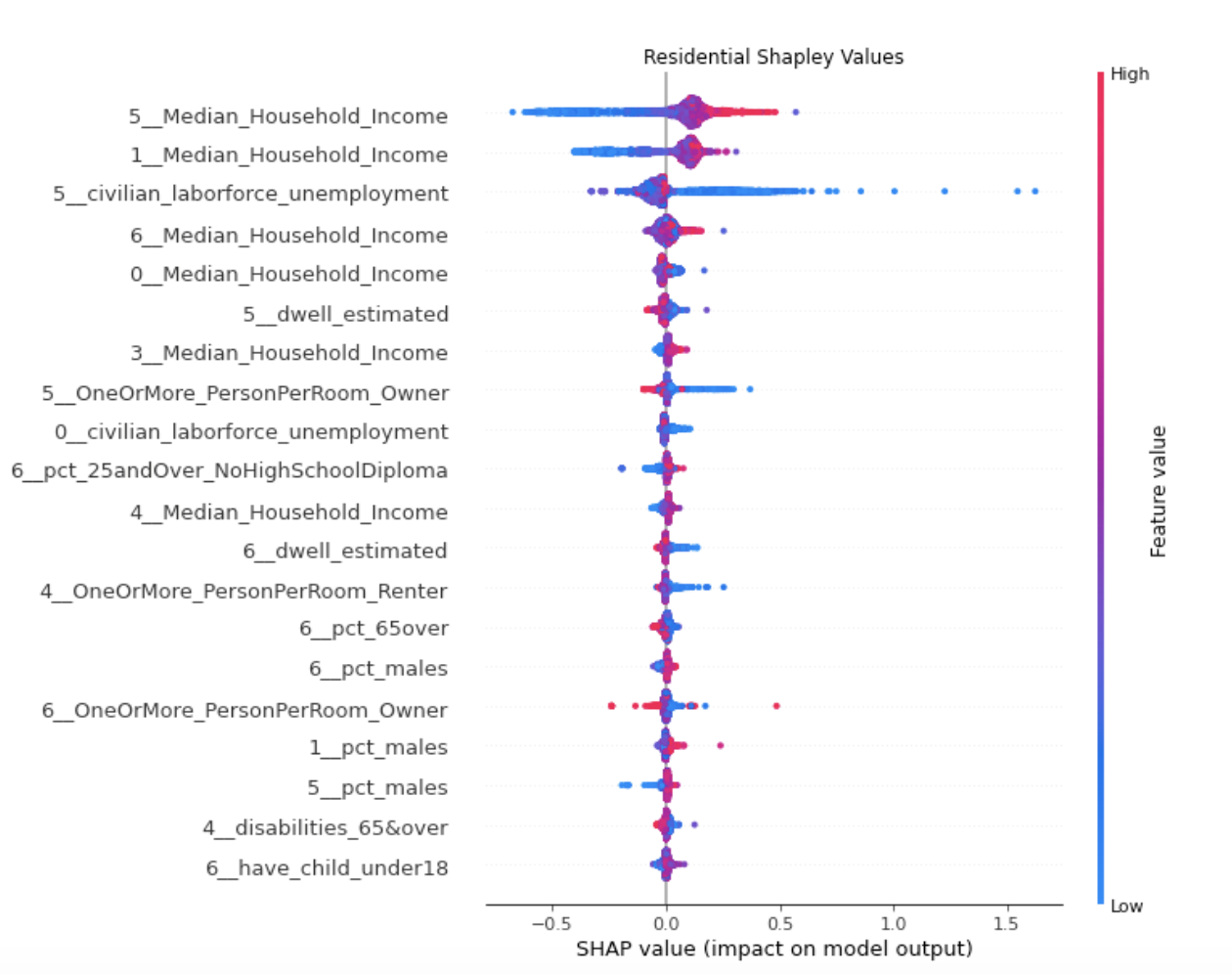}
\end{figure}

A few key differences emerge when comparing the commercial and residential shapley values. Firstly, commercial properties seem to be much more impacted by the number of people in the area than residential. In particular, this feature is important on weekdays (Tuesday and Wednesday) for commercial properties. This impact could be driven by the fact that commercial real estate such as office buildings derive much of their value by how many people go to work there during the week. 
Additionally, the proportion of people in the area that are commuting tends to impact commercial more than residential. Interestingly, this relationship is negative during the week (on Tuesday) and positive on the weekend (on Sunday). One possible reason for this is that during the week if a smaller number of people are stopping in an area, it may not be a popular work destination (and thus less valuable). However, on Sundays when people are less likely to work, a higher proportion of people moving in an area could mean it is a nice area to walk in. 
Another major difference between the two types of properties is that the estimated median household income of non-residents in an area is much more impactful for residential properties. Unsurprisingly, people are willing to spend more on homes which have more affluent people traveling in the area even if they do not live there. Together these differences show that mobility data does in fact impact commercial and residential properties differently.

\section{Discussion}
Based on the results of our analysis, mobile location data seems to be beneficial when creating real estate models. Though the increase in performance was relatively small, more advanced models could possibly make this increase even more pronounced. Furthermore, we were able to identify key differences between mobility’s relationship with residential and commercial properties. While median income seemed to be the primary driver of residential value, the number of people in the area had the largest impact on commercial value. 

\section{Conclusion}
In conclusion, the inclusion of mobile location data can help machine learning models more accurately predict residential real estate prices. We made our final model with our dynamic mobile location features, and then created the exact same model but without the dynamic mobile location features; the model with the dynamic mobile location features had a lower mean squared error. Hence, the inclusion of the dynamic mobile location features did in fact improve the model. One of our main takeaways is that stacking two models, one that only looks at static features and one that looks at only dynamic features gets the most benefit from the inclusion of the mobile location features. In terms of future work,  attempting different types of stacked models is a good next step. Additionally, we struggled to compare how the mobile location data impacted the residential versus commercial real estate data. This comparison would be easier with more commercial real estate features.

\section*{Acknowledgments}
We would like to thank the UVA Data Science Institute, and the
UVA School of Commerce. In particular, we would like to
acknowledge Dr. Heman Shakeri and Dr. Natasha Zhang Foutz. Their expertise, guidance, and support made this study possible.

\section*{Author Information}
\noindent\textbf{Walter Coleman,} M.S. Student, Data Science Institute, University of Virginia \\
\textbf{Ben Johann,} M.S. Student, School of Data Science, University of Virginia \\ 
\textbf{Nicholas Pasternak,} M.S. Student, School of Data Science, University of Virginia \\
\textbf{Jaya Vellayan,} M.S. Student, School of Data Science, University of Virginia \\
\textbf{Natasha Foutz,} Associate Professor, McIntire School of Commerce, University of Virginia \\
\textbf{Heman Shakeri,} Assistant Professor, School of Data Science, University of Virginia \\

\vspace{12pt}

\end{document}